# Deep Learning Pipeline for Preprocessing and Segmenting Cardiac Magnetic Resonance of Single Ventricle Patients from an Image Registry

Tina Yao, MRes* • Nicole St. Clair, BSc* • Gabriel F. Miller, MSc • Adam L. Dorfman, MD • Mark A. Fogel MD • Sunil Ghelani, MD • Rajesh Krishnamurthy, MD • Christopher Z. Lam, MD • Joshua D. Robinson, MD • David Schidlow, MD • Timothy C. Slesnick, MD • Justin Weigand, MD • Michael Quail, MD • Rahul Rathod, MD • Jennifer A. Steeden, PhD • Vivek Muthurangu, MD

* T.Y. and N.St C. contributed equally to this work

From the Institute of Health Informatics (T.Y.), Institute of Cardiovascular Science (J.S., V.M.), University College London, London, England; Department of Cardiology, Boston Children's Hospital, MA, USA (N.St C., G.F., S.G., D.S., R.R.); Pediatrics and Communicable Disease, University of Michigan, Ann Arbo, MI, USA (A.D.); Division of Cardiology, The Children's Hospital of Philadelphia, PA, USA (M.F.); Radiology, Nationwide Children's Hospital, OH, USA (R.K.); Diagnositic Imaging, Hospital for Sick Children, Toronto, Canada (C.L.); Department of Pediatrics, Ann and Robert H Lurie Children's Hospital of Chicago, IL, USA (J.R.), Pediatric Cardiology, Children's Healthcare of Atlanta Inc (T.S.), Department of Cardiology, Texas Children's Hospital (J.W.).

**Address correspondence to**: V.M. (email: v.muthurangu@ucl.ac.uk)

***Abbreviations (10):*** DL = deep learning, CMR = Cardiac Magnetic Resonance, SAX = short axis, CNN = convolutional neural network, EDV = end-diastolic volume, ESV = end-systolic volume, SV = stroke volume, EF = ejection fraction

***Summary Statement***

An end-to-end deep learning pipeline was developed to provide automatic segmentation and cardiac function metrics for a registry of single ventricle patients. The pipeline requires no human input and is the first to segment single ventricle patients.

***Key Points***

- Deep learning segmentation pipeline can provide automated 'core-lab' processing of a registry of single ventricle patients that is robust to highly variable anatomy and heterogeneous data collected from >10 hospitals
- The pipeline achieved a median Dice score of 0.91 (0.89-0.94) for end-diastolic volume, 0.86 (0.82-0.89) for end-systolic volume, and 0.74 (0.70-0.77) for myocardial mass. The deep learning derived end-diastolic volume, end-systolic volume, myocardial mass, stroke volume and ejection fraction had no statistical difference compared to the same values derived from manual segmentation with p-values all >0.05
- The pipeline was further tested on 475 unseen patient exams, achieving 68% adequate segmentation in both systole and diastole, 26% needed minor adjustments in either systole or diastole, 5% needed major adjustments, and the cropping model only failed in 0.4%

## Abstract

***Purpose:*** To develop and evaluate an end-to-end deep learning pipeline for automated segmentation and analysis of cardiac magnetic resonance images to provide core-lab processing for a multi-centre registry of Fontan patients.

***Materials and Methods:*** This retrospective study used training (n = 175), validation (n = 25) and testing (n = 50) cardiac magnetic resonance image exams collected from 13 institutions in the UK, US and Canada. The data was used to train and evaluate a pipeline containing three deep learning models, including a CNN classifier to identify short-axis cine stacks, a UNet 3+ model to find and crop the heart from the image and another UNet 3+ model to segment the blood pool and myocardium. The pipeline's performance was assessed on the Dice and IoU score between the automated and reference standard manual segmentation. Cardiac function values were calculated from both the automated and manual segmentation and evaluated using Bland-Altman analysis and paired t-tests.

The overall pipeline was further tested on 475 unseen patient exams, and the predicted segmentation results were qualitatively evaluated on whether the segmentation was adequate or needed adjustment.

***Results:*** For the 50 testing dataset, the deep learning pipeline achieved a median Dice score of 0.91 (0.89-0.94) for end-diastolic volume, 0.86 (0.82-0.89) for end-systolic volume, and 0.74 (0.70-0.77) for myocardial mass. The deep learning derived end-diastolic volume, end-systolic volume, myocardial mass, stroke volume and ejection fraction had no statistical difference compared to the same values derived from manual segmentation with p values all greater than 0.01.

For the 475 unseen patient exams, the pipeline achieved 68% adequate segmentation in both systole and diastole, 26% needed minor adjustments in either systole or diastole, 5% needed major adjustments, and the cropping model only failed in 0.4%.

***Conclusion:*** Automated deep learning pipelines can provide standardised 'core-lab' segmentation for single ventricle patients across multiple centres. This pipeline can now be applied to the >4500 cardiac magnetic resonance exams currently in the FORCE registry as well as any new patients that are recruited.

# Introduction

Approximately 4-8 in 10000 newborns are born with a functional single ventricle, and most will undergo a total cavopulmonary connection, leaving them with a Fontan circulation (1). These patients are at risk of heart failure, and cardiovascular magnetic resonance (CMR) is considered the reference standard method of evaluating ventricular size and function. Several small studies have shown that CMR metrics of ventricular function are predictive of outcomes (2), but larger studies are needed to truly understand the importance of CMR data in this patient group (3).

The Fontan Outcomes Registry Using CMR Examinations (FORCE) is the first large-scale (>4500 CMR scans in >3000 unique patients), multi-centre CMR registry of patients who have undergone Fontan palliation (4). Although quantitative ventricular volume data from each site is included in the registry, significant differences in segmentation protocols and interobserver variability make this data difficult to use. As image data is part of FORCE, 'core-lab' processing is possible, but manual segmentation of the whole registry is not practical or feasible. Thus, automated methods are needed to fully harness the potential of FORCE.

Recent innovations in deep learning (DL) have enabled automated cardiac segmentation methods to reach human levels of accuracy (5–8). However, most DL models are trained and validated on structurally normal hearts (9–11), and automated segmentation in congenital heart disease (CHD) poses a significantly greater challenge (12,13). Deep learning segmentation models have been successfully developed for biventricular CHD, but none are currently suitable for functionally single ventricles (14,15).

Therefore, we propose a DL pipeline trained on data from the FORCE registry that automatically identifies ventricular short axis (SAX) cine stacks, crops out the heart and segments the ventricles of Fontan patients. The pipeline approach is vital for use in registries as no human input is required. Recently, Govil et al. have demonstrated that such an end-to-end pipeline is feasible for biventricular CHD (16), and our method extends this to more complex single ventricular anatomy from a registry that currently contains data from >15 sites.

The aims of this study were to: i) develop and validate each section of the pipeline on a curated dataset with "ground truth" segmentations, ii) test the pipeline on large amounts of registry data, and iii) compare the site entered and DL derived volumetric data.

# Methods

## Pipeline Overview

The automated pipeline (Figure 1) consists of 4 stages: (1) cine stack extraction, (2) SAX identification, (3) heart localisation and cropping, (4) ventricular segmentation and derivation of clinical values.

## Training Dataset Preparation

The training dataset for all DL models (used in stages 2-4) consisted of complete CMR exams of 250 patients from the FORCE registry. This retrospective study was approved by each institution's Committee on Clinical Investigation through a separate application or via reliance agreement, with all images de-identified on upload. Patients were scanned at 13 institutions across three countries between November 2007 and December 2022, at both 1.5 and 3.0T, using three MRI manufacturers (demographics in Table 1). The dataset was split 175/25/50 for training, validation, and testing. The training patients were stratified such that the proportion of patients from each site was the same as in the full database. However, for validation and testing, the number of patients from each hospital was kept approximately the same.

For each patient exam, the SAX data were segmented by a clinical researcher (3.5 years cardiac imaging experience) and later reviewed and adjusted by three cardiovascular imaging physicians (6, 8, 14 years experience, respectively). Endocardial and epicardial contours at end-systole and end-diastole were manually traced (Circle cvi42 version 5.14.2, Circle Cardiovascular Imaging, Calgary, Alberta, Canada) as per standard convention, with trabeculae and papillary muscles included in the blood pool. In patients with underdeveloped left or right second ventricles, both ventricles were contoured. For training of DL networks, traced contours were converted into two binary masks for the blood pool (combined if two ventricles were present) and myocardium. This ground truth data was treated as 'core-lab data' as it was segmented by one person, but each patient exam also had calculated volumes and mass from their host sites that were segmented by numerous people using different protocols (site data). This allowed comparison of the core-lab and site-entered volumetric data.

## Stage 1: Cine Stack Extraction

Extraction of all the cine stacks in each patient study was achieved by using information in the DICOM headers (17). Specifically, 2D cine data (determined by series with at least 10 frames in the same slice location) were designated to belong to the same stack (in any orientation) if the images had the same orientation (Image Orientation Patient Attribute) and pixel size (Pixel Spacing Attribute) separated by their space between slices (Spacing Between Slices Attribute). Cine data had to contain at least six slices to be considered a stack.

### Stage 2: Short-Axis Identification

A model was trained to select the SAX cine stack from other cine stacks acquired in a single exam (e.g. 4-chamber, transverse or long axis stacks). The training data consisted of the first phase (assumed to be diastolic) of the central five slices of all the cine stacks in the 175 training exams, labelled either SAX or non-SAX. Non-square images were zero-padded to their largest side, and then all images were resized to 128x128 pixels. The classifier was a CNN followed by two fully connected dense layers and a sigmoid final layer that outputs the probability that the input image is in the SAX orientation ($P_{sax}$), trained with a binary cross-entropy loss (see supplementary information for full architecture and training details).

At inference, the first phase of the central five slices of each cine stack was inputted into the classifier. The stack that contained the slice with the highest $P_{sax}$ was identified as the SAX orientation. If more than one stack had the same maximum probability, then the stack with the highest mean $P_{sax}$ was chosen.
For testing (50 exams), accuracy, precision and recall were assessed per image using a threshold of $P_{sax} > 0.5$ for SAX identification. In addition, the ability to correctly identify the SAX stack per exam was assessed using the process described above.

### Stage 3: Heart Localisation and cropping

Cropping the heart is necessary for a robust pipeline as it centers the heart regardless of its size or position in the original image, thereby producing more accurate segmentation results (18). Heart localisation necessary to crop the region of interest was reframed as a simplified segmentation problem. The model was trained to predict ground truth binary 'whole heart' masks based on the manually segmented epicardial border. The model was trained using the segmented end-diastolic frames of all slices in the SAX stacks of the 175 training exams. Non-square images were zero-padded to the largest side and then all images were resized to 256x256 pixels.

The segmentation model was based on a modified UNet 3+ architecture, an improvement on the conventional UNet model that utilises full-scale skip connections and can use deep supervision (19). The hyperparameter-optimized UNet 3+ was trained using Intersection over Union (IoU) loss with deep supervision (see supplementary information for full architecture and training details).

The method used for cropping is illustrated in Figure 2. At inference, the model predicted the 'whole heart' mask in the first (assumed to be diastolic) phase for all slices in the SAX stack. Any disconnected 'islands' in the predicted masks were removed if they did not overlap with the intersection of the masks in the slice direction. This effectively removes common misclassified regions such as the stomach. Both the predicted and ground truth bounding boxes were calculated from the segmentation masks using the same method. First, a preliminary bounding box was defined as the minimum square containing the union of all the segmentation masks. Then, the bounding box was expanded by 50% for redundancy. The box is used to crop the location of the heart through all slices.

For testing (50 exams), the predicted and ground truth bounding boxes were compared using IoU.

### Stage 4: Ventricle Segmentation

The final segmentation model used the UNet 3+ architecture and was trained to predict three pixel classes: blood pool, myocardium and background. The model was trained on all the end-diastolic and end-systolic slices in the SAX stacks of the 175 training exams. Images that did not contain ventricular structures were labelled as all background. All images were cropped using bounding boxes created from the ground truth masks as described above and resized to 128x128 pixels.

The hyperparameter optimized UNet 3+ was trained using Tversky loss, but no deep supervision (see supplementary information for full architecture and training details).

At inference, all slices and frames of the SAX stack were segmented. Post-processing included removing any component separate from the heart, where the heart is defined as the largest connected component in 3D. From these masks, ventricular volume-time curves were created by summing the labelled blood pool voxels (scaled by voxel volume) across all slices for all cardiac phases. End systole and diastole were identified as the phases with the smallest and largest volumes using the volume time curve constrained to the middle five slices (to lessen any effect of poor segmentation at the base or apex). End systolic and diastolic volumes (ESV and EDV) were then defined as the volume (assessed over all slices) at these timepoints. Stroke volume (SV) was EDV – ESV, and ejection fraction (EF) was SV divided by EDV. Myocardial mass was the sum of the myocardial voxels (multiplied by myocardial density) in the end-diastolic phase. All volumes and mass were indexed to body surface area.

For testing (50 exams), the predicted and ground truth blood pool and myocardial masks were compared per slice using IoU and Dice for end-diastole and end-systole. In addition, ground truth and predicted ESV, EDV, SV and EF were compared using Bland Altman and Correlation analysis.

### Pipeline Performance

The function of the whole pipeline was tested by processing 475 totally new exams (not used for previous training or testing) - including data from three sites not represented in the training data (demographics in Table 1). All end-diastolic and end-systolic segmentations were examined by the clinical researcher, and the pipeline results were rated as follows (see supplementary information for full details): i) crop failure (part or all of the heart missing from cropped images), ii) satisfactory segmentation (appropriate for use in a clinical context), iii) segmentation requiring minor adjustments (small adjustment required in 1-2 slices), and iv) segmentation requiring major adjustment (significant failure in a majority of slices). In addition, subjective image quality was rated as satisfactory or sub-optimal. In those studies deemed to have satisfactory segmentation or only requiring minor adjustment, volumetric data was compared to data entered by the host site (using varying clinical post-processing protocols).

### Statistics

Continuous variables are expressed as median (interquartile range). Bland-Altman and correlation analysis were used to assess agreement between DL, manual ventricular segmentation and site-entered data. The biases between manual and DL segmentation were compared using a paired t-test. Comparison of pipeline results for different ventricular types, paediatric vs. adult and 1.5T vs 3T was performed using the Chi-squared test. Statistical analyses were performed in R, a p-value of less than 0.05 indicated a significant difference.

## Results

### Short-axis Identification Classifier and Heart Localisation Model

The accuracy for SAX identification per slice was 96.1%, precision was 98.0%, and recall was 94.4%. However, because all slices are evaluated to make a final decision, the classifier was able to correctly identify the short-axis stack in all 50 test exams.

For heart localisation, the mean IoU between the ground truth and predicted bounding boxes per exam was 0.94 (0.92-0.96). More importantly, the calculated bounding box contained the whole heart for all 50 test exams.

### Ventricle Segmentation Model

Figure 3 shows examples of ventricle segmentation through the pipeline for the best, median and worst cases (regardless of systolic or diastolic phase), based on Dice score (movies of all frames in supplementary information).

There was acceptable agreement between the automated DL segmentation and expert manual segmentation for both EDV and ESV with no significant biases ($p>0.56$), acceptable limits of agreement and high levels of correlation (Figure 4). This also resulted in acceptable agreement with no significant biases ($p>0.51$) for SV and EF (Figure 5). In addition, there was moderate agreement ($p=0.12$) for ventricular mass (Figure 5). The Dice score was 0.91 (0.89-0.94) and 0.86 (0.82-0.89) for EDV and ESV, respectively, while the IoU Score was 0.84 (0.80-0.88) and 0.76 (0.70-0.81) for EDV and ESV. The myocardial mass had a Dice score of 0.74 (0.70-0.77) and an IoU score of 0.59 (0.54-0.63).

### Pipeline Performance

Pipeline processing was feasible in all 475 new cases, with all SAX stacks correctly identified. Table 2 shows the time taken and the number of images processed at each stage. The average time to process a patient exam through the pipeline was 26s (IQR: 21 - 32s).

Of the 475 patients, 68% had satisfactory segmentation of both their end-systolic and end-diastolic volumes, 26% needed minor adjustments in at least one volume, 5% needed major adjustments, and in 0.4% the cropping model failed (see supplementary information for definition and examples). Of the total 950 volumes, 80% had satisfactory segmentation quality (end-systole 78%, end-diastole 82%). Approximately 35% of images were identified as having suboptimal image quality, and there was a statistically lower ($p<0.0001$) proportion of satisfactory segmentations in this group (Table 3). Segmentation was also more successful in 1.5T images compared to 3.0T ($p=0.00088$). There was no significant difference between segmentation success for different ventricle types ($p=0.11$), different vendors ($p=0.13$), or paediatric vs adult patients ($p=0.36$).

There was only moderate agreement between DL-generated volumes and those entered by the host site (Figure 6), however, this level of agreement was similar to that between the core-lab manually segmented and the host site volumes (Figure 6).

## Discussion

This study is the first description of DL automated segmentation of single ventricles. The main findings of this study were: i) It is feasible to create an end-to-end deep learning pipeline that automatically takes CMR exams, extracts SAX cines, performs cropping, and segments the ventricles, ii) There was acceptable agreement between DL and manual segmentation of functional single ventricles in terms of volumes and mass, iii) The pipeline can rapidly and accurately segment large numbers of unseen cases with a high degree of success and iv) There was only moderate agreement between site entered volumes, and both DL and manually segmented core-lab data. Unlike previous segmentation pipelines, our framework processes images straight from an image registry (16) and could process the whole FORCE registry in <40 hours.

Although the use of DL for ventricular segmentation in CMR is well described, its use in CHD is much more limited, possibly due to less access to large training datasets and more complex anatomy (13).

Nevertheless, successful models have been produced (12,14–16) for biventricular CHD, including pipeline frameworks for Tetralogy of Fallot (14). However, creating a pipeline that is capable of segmenting extremely heterogeneous single ventricular anatomy from a large number of institutions with a wide range of protocols, scanners, and field strengths is significantly more difficult. Thankfully, the size of the FORCE registry allowed creation of a large training dataset collected from 13 different hospitals, aiding generalizability. Furthermore, we choose to use the UNet 3+, which has been shown to improve segmentation accuracy over more simple UNet architectures.

Our DL segmentation pipeline demonstrated acceptable agreement with manual segmentation for ventricular volume, mass, and ejection fraction, with results comparable to other DL models for CMR segmentation in CHD patients (12,14–16). Furthermore, our DL always produces the same result, which is highly desirable, as it has been shown that there is significant interobserver variability of CMR metrics in Fontan patients (20).

More importantly, when applied to the FORCE registry (from 16 institutions, including data from 3 sites not included in training), our end-to-end pipeline was successful in 68% of cases, with 32% of exams (20% of volumes) requiring some adjustment. When applied across the whole FORCE registry, this would represent ~3000 exams processed without any user input. Further improvements could also be achieved by fine-tuning the model with manual segmentation of images where the DL model has failed. Such an approach would benefit from automated quality assurance (QA), which could determine segmentation accuracy (21,22) and automatically identify data that requires resegmentation. Several groups have developed automated QA for ventricular segmentation, and we could use the segmentation review data from the 475 datasets that underwent full pipeline processing as training data.

It should be noted that the agreement between the DL-derived volumes and the site-entered volumes was only moderate. However, this was also the case for core-lab manually segmented volumes and the site-entered data, suggesting any lack of agreement is mainly due to institution-specific differences in segmentation and interobserver variability. This further demonstrates the importance of core-lab processing, particularly when comparing institutions or evaluating longitudinal changes.

The main limitation of this study was the use of manual segmentations from a single operator. This was done to ensure consistency of segmentation but could result in biases in the model. However, all segmentations were reviewed and adjusted by one of three experts, which we believe minimises any problematic biases.

To conclude, we have demonstrated a scalable pipeline for automated segmentation of ventricular volumes for CMR scans of Fontan patients. Importantly, our pipeline is fully integrated into the largest imaging registry in this patient population. We believe that combining the registry's clinical data with these automated cardiac function metrics will give researchers and clinicians new insights into the role of CMR in the management of the Fontan circulation.

## Acknowledgements

**Funding**: This work was supported through a grant from the Additional Ventures Foundation

# Figure Legends

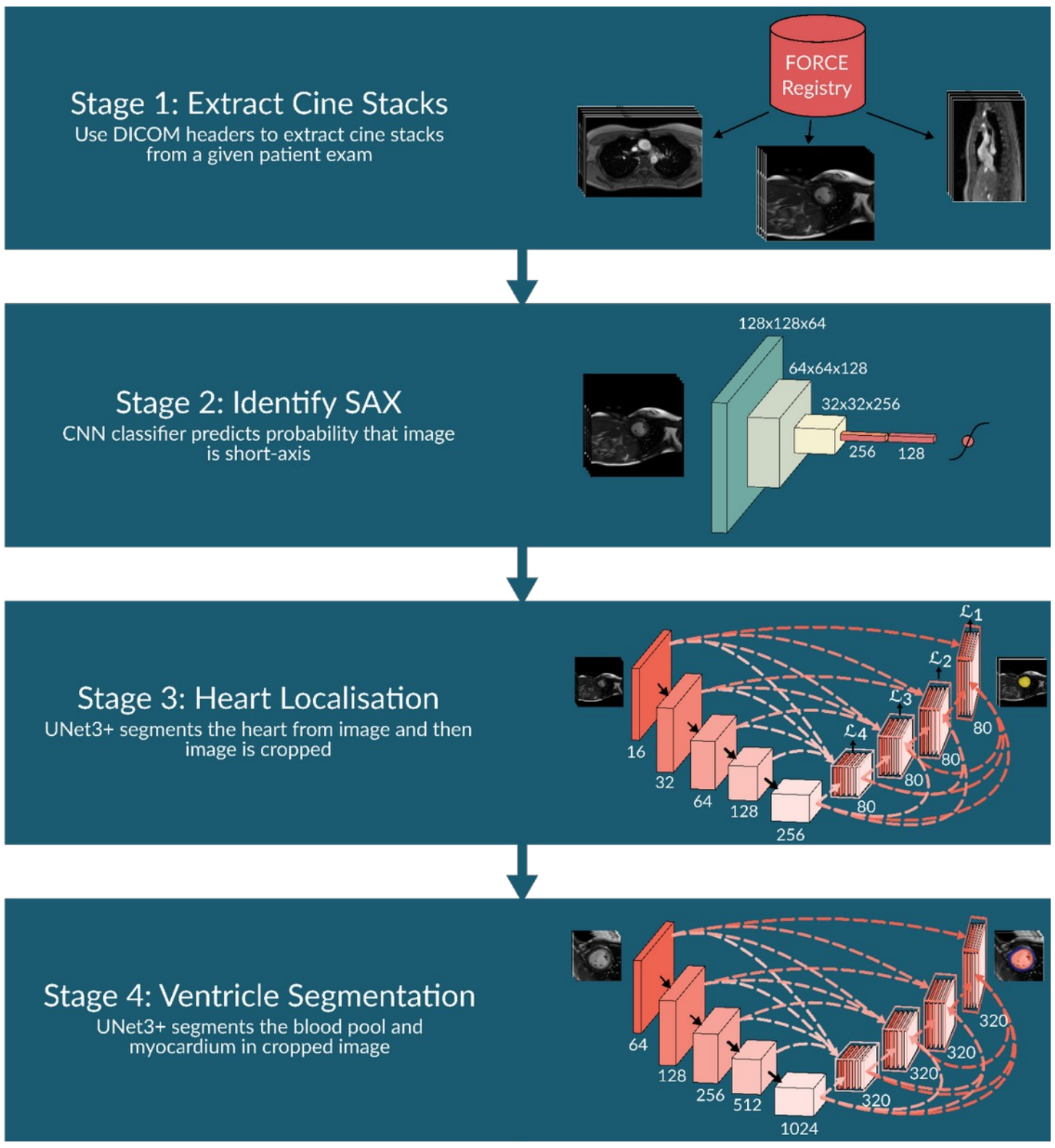

Figure 1. Outline of the 4 stages involved in the deep learning pipeline.

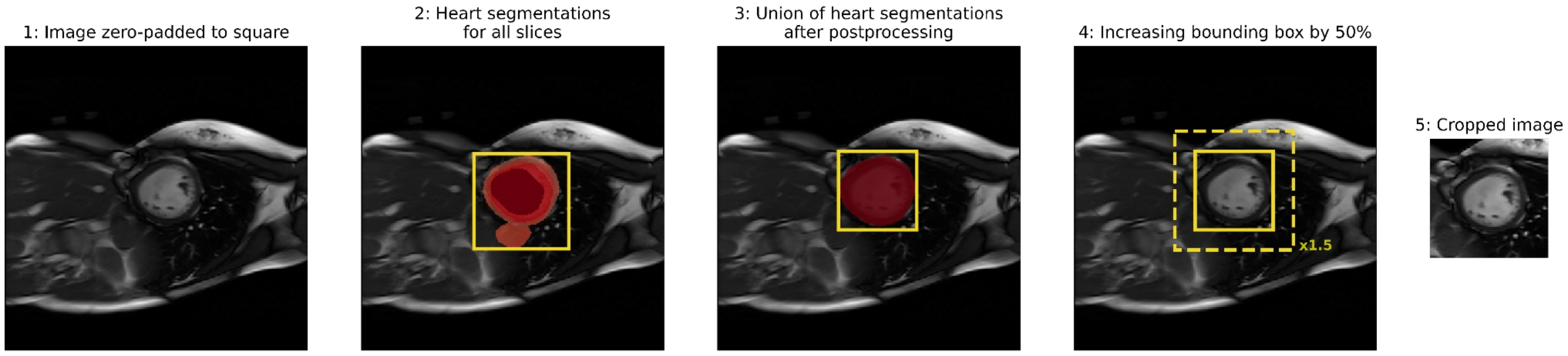

Figure 2. Visual example of the steps involved in heart localisation. Red shows the segmentation prediction for the heart and the yellow box is the bounding box. Step 2 shows the segmentation for all slices, where the darker the red, means greater agreement between slices. Post-processing in step 3 removes any heart segmentation, for a given slice, that does not overlap with the intersection of all the heart segmentations in step 2 (dark red region).

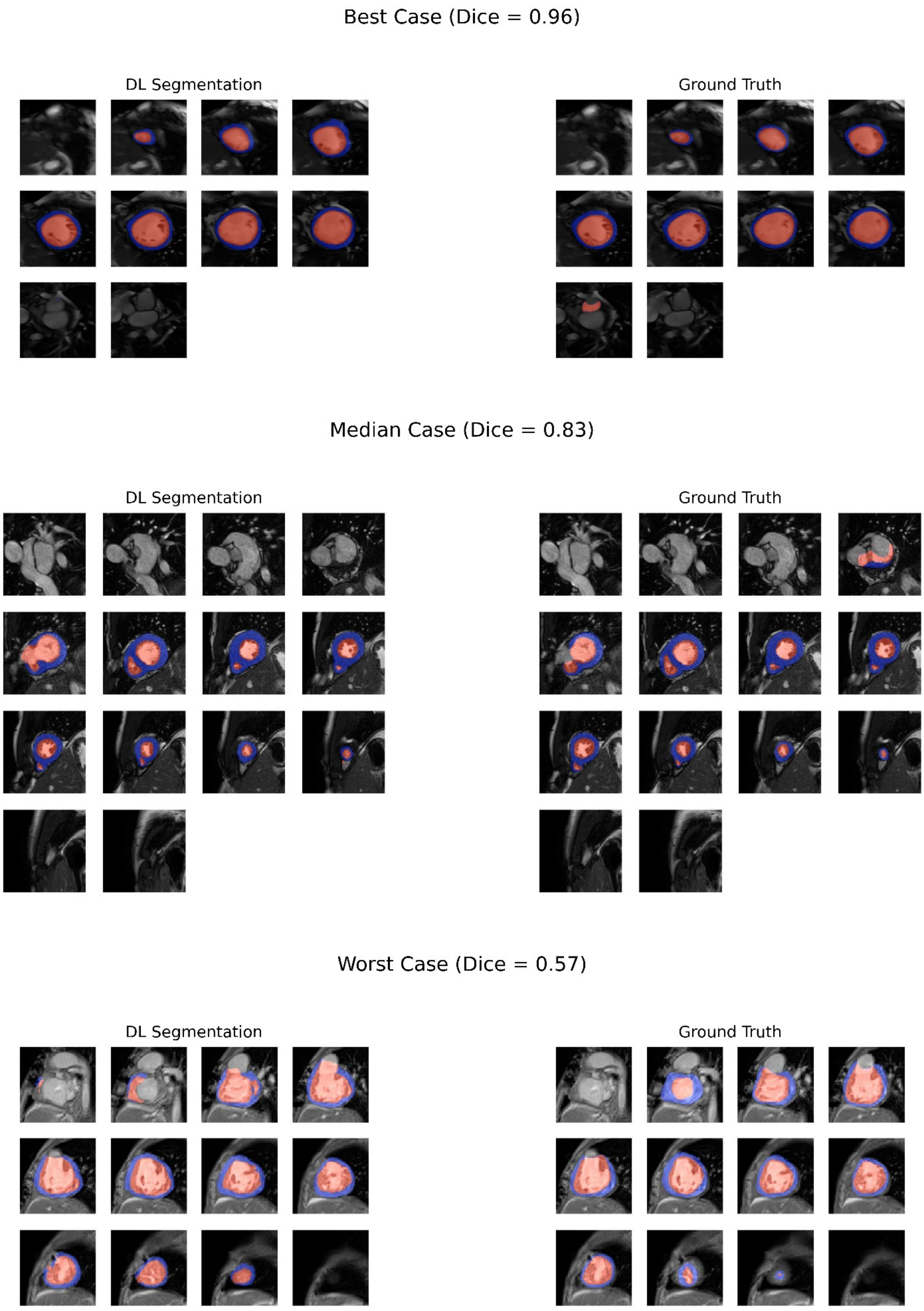

Figure 3. Machine learning segmentations vs ground truth segmentations for the best, median and worst test cases. Blood pool is shown in red and myocardium in blue.

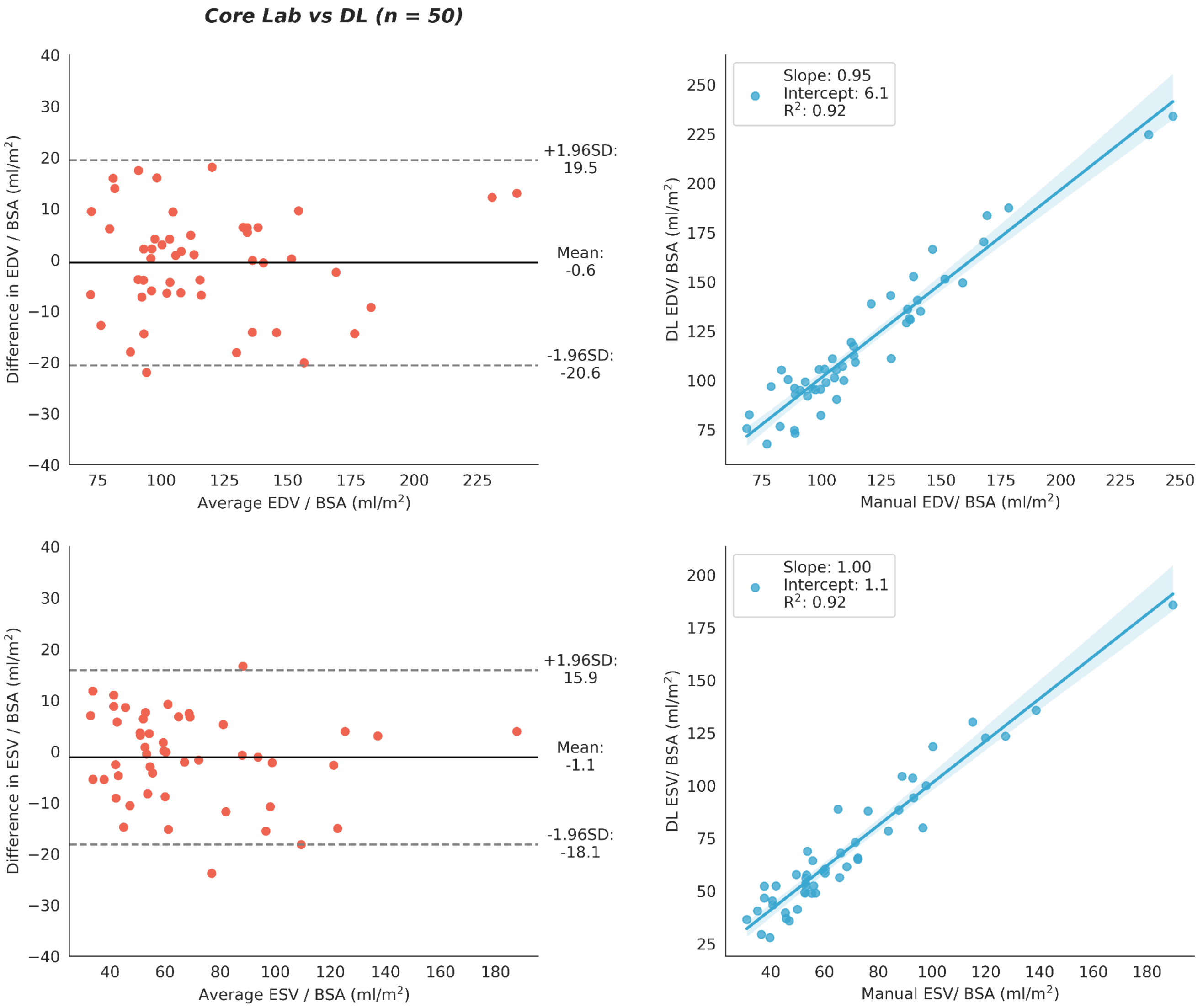

Figure 4. Bland-Altman and correlation plot comparing the Manually-derived vs DL-derived end-diastolic and end-systolic volumes.

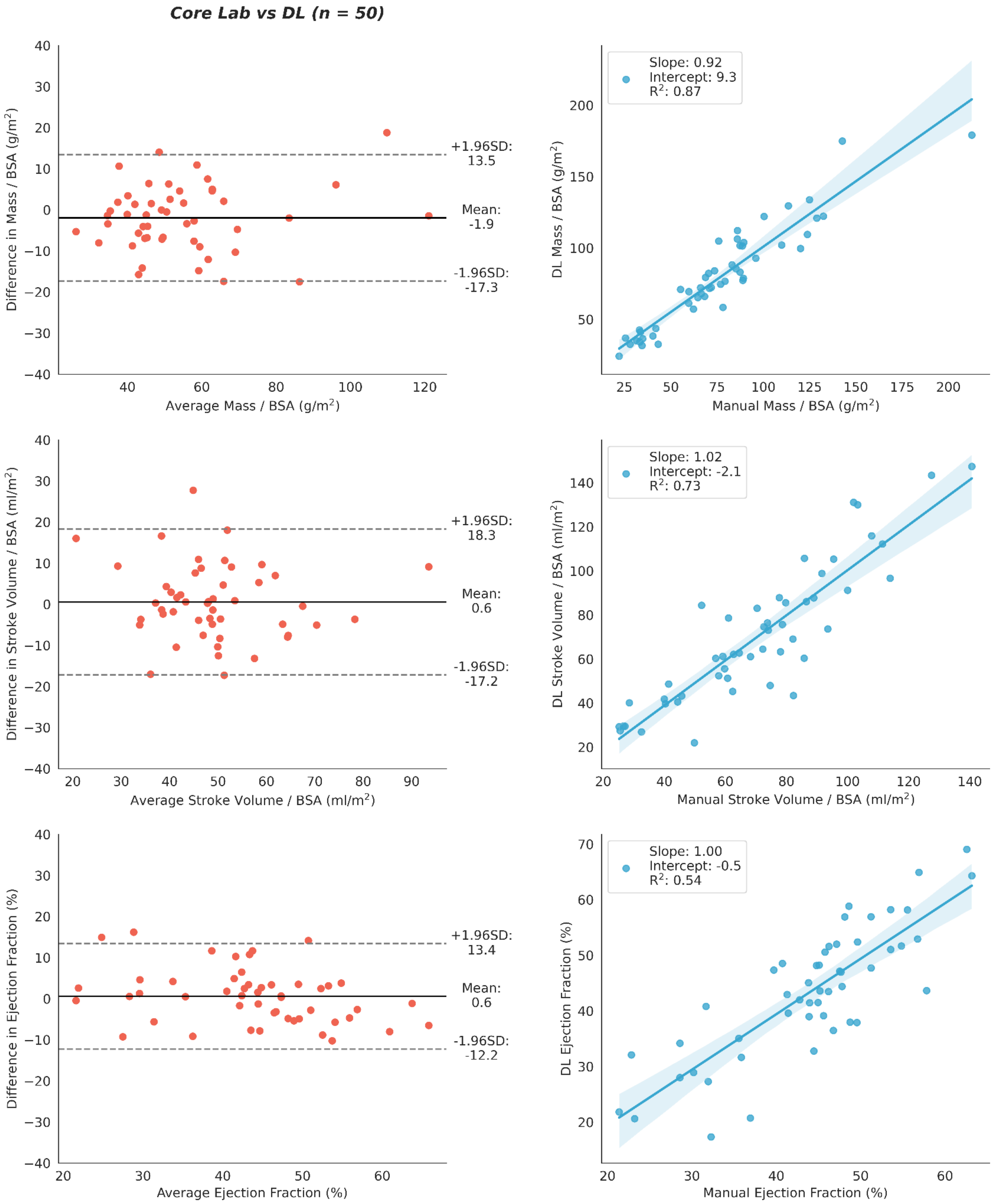

Figure 5. Bland-Altman and correlation plot comparing the Manually-derived vs DL-derived ventricular mass, stroke volume and ejection fraction.

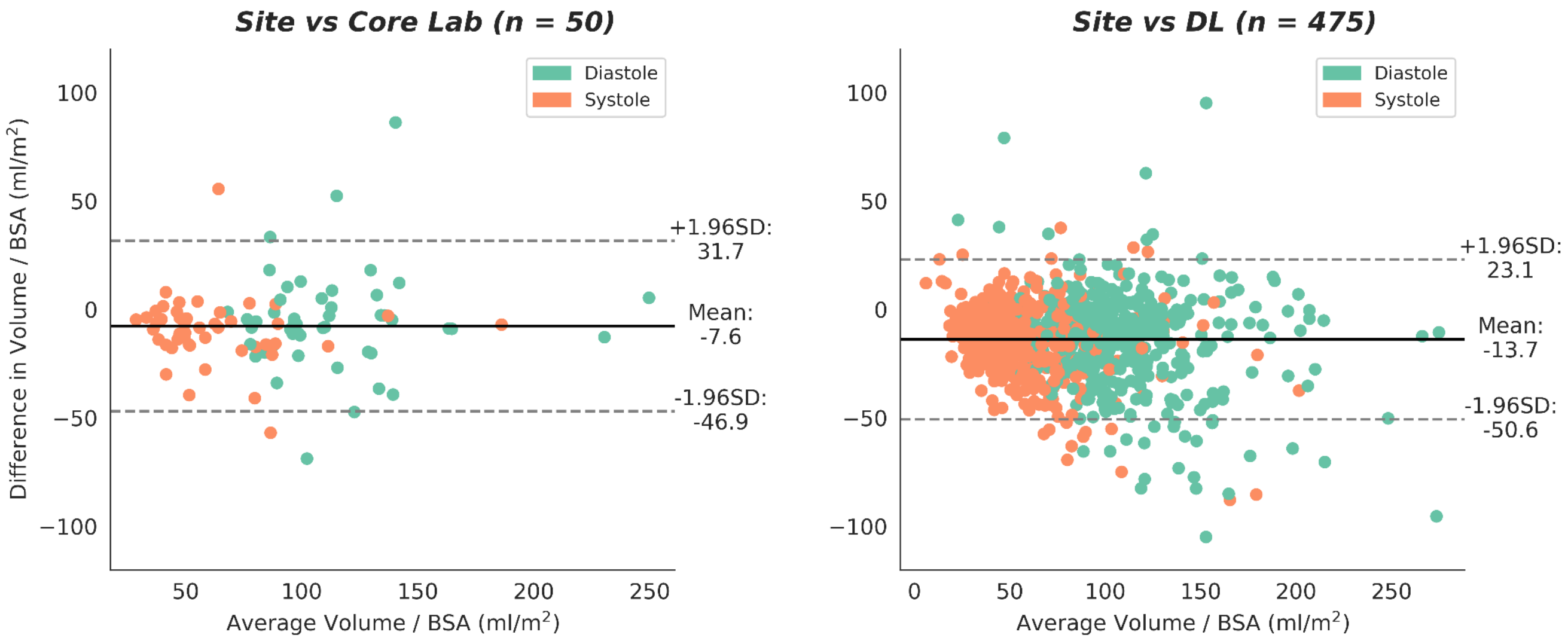

Figure 6. Bland-Altman for ventricular volume. (left) Comparison between site-entered volume vs manually-derived volume for 50 test patients. (right) Comparison between site-entered volume vs DL-derived volume for 475 test patients.

# Tables

## Table 1

| | | Number of patients in the manually segmented training set (%)<br>*n = 250* | Number of patients in the pipeline test set (%)<br>*n = 475* |
|---|---|---|---|
| Age | | | |
| | Adult | 137 (55%) | 212 (45%) |
| | Child (<16 years) | 113 (45%) | 263 (55%) |
| Ventricle Systemic Circulation | | | |
| | Both | 133 (55%) | 238 (50%) |
| | RV Only | 64 (26%) | 137 (29%) |
| | LV Only | 47 (19%) | 100 (21%) |
| Scanner Field Strength | | | |
| | 1.5 T | 237 (95%) | 458 (96%) |
| | 3.0 T | 13 (5%) | 17 (4%) |

Table 1: Demographics of the 250 patient exams used for training and validation of the deep learning models and of the 475 patients used for testing the deep learning pipeline.

## Table 2

| Stage | Variable | Median (Interquartile Range) | Range |
|---|---|---|---|
| 1: Extract Cine Stacks | | | |
| | Number of files | 3300 (1900 - 7100) | 460 - 21,000 |
| | Number of series | 35 (23 - 58) | 5 - 290 |
| | Number of stacks | 2 (1 - 3) | 1 - 16 |
| | Time Taken (s) | 9 (5 - 12) | 1 - 73 |
| 2: Identify SAX | | | |
| | Number of images classified | 10 (5 - 15) | 5 - 80 |
| | Time Taken (s) | 0.40 (0.20 - 0.48) | 0.030 - 1.45 |
| | Time per image (s) | 0.080 (0.040 - 0.096) | 0.0060 - 0.2900 |
| 3: Heart Localisation | | | |
| | Number of images used for localisation | 12 (12 - 14) | 7 - 25 |
| | Time Taken (s) | 1.5 (1.4 - 1.7) | 1.2 - 5.7 |
| | Time per image (s) | 0.12 (0.11 - 0.13) | 0.073 - 2.200 |
| 4: Ventricle Segmentation | | | |
| | Number of images segmented | 340 (240 - 360) | 100 - 1600 |
| | Time taken (s) | 11 (9 - 12) | 4 - 61 |
| | Time per image (s) | 0.032 (0.031 - 0.034) | 0.028 - 0.068 |
| Total time taken (s) | | 26 (21 - 32) | 13 - 110 |

Table 2: Median, interquartile range and range of the time taken and number of images processed for each stage of the pipeline for the 475 test set, using one NVIDIA GeForce RTX 3090 GPU. Values are rounded to 2 significant figures.

## Table 3

| Contains Artefact and/or has Poor Image Quality | | Adequate | Minor Adjustments | Major Adjustments | Crop Fail |
|---|---|---|---|---|---|
| n = 475 | | | | | |
| All | 475 | 323 (68%) | 124 (26%) | 26 (5%) | 2 (0.4%) |
| No | 311 | 230 (74%) | 73 (23%) | 8 (3%) | 0 (0%) |
| Yes | 164 | 93 (57%) | 51 (31%) | 18 (11%) | 2 (1%) |

Table 3: Labels given to the end-diastolic and end-systolic segmentation output of the pipeline for the 475 patients.